\pdfoutput=1

\documentclass{article}
\usepackage{tenor2020}
\usepackage{ifpdf}
\usepackage[english]{babel}
\usepackage{balance}
\usepackage[htt]{hyphenat}

\def\papertitle{PAPER TEMPLATE FOR TENOR 2020}
\def\firstauthor{First author}
\def\secondauthor{Second author}
\def\thirdauthor{Third author}

\newif\ifpdf
\ifx\pdfoutput\relax
\else
   \ifcase\pdfoutput
      \pdffalse	
   \else
      \pdftrue
\fi

\ifpdf 
  \usepackage[pdftex,
    pdftitle={\papertitle},
    pdfauthor={\firstauthor, \secondauthor, \thirdauthor},
    bookmarksnumbered, 
    pdfstartview=XYZ 
   ]{hyperref}

  \usepackage[pdftex]{graphicx}
  \graphicspath{{./figures/}}
  \DeclareGraphicsExtensions{.pdf,.jpeg,.png}

  \usepackage[figure,table]{hypcap}

\else 
  \usepackage[dvips,
    bookmarksnumbered, 
    pdfstartview=XYZ 
  ]{hyperref}  

  \usepackage[dvips]{epsfig,graphicx}
  \graphicspath{{./figures/}}
  \DeclareGraphicsExtensions{.eps}

  \usepackage[figure,table]{hypcap}
\fi

\hypersetup{
    colorlinks,%
    citecolor=black,%
    filecolor=black,%
    linkcolor=black,%
    urlcolor=black
}
\usepackage[center]{caption}
\usepackage{verbatim}
\usepackage{tikz}
\def\checkmark{\tikz\fill[scale=0.4](0,.35) -- (.25,0) -- (1,.7) -- (.25,.15) -- cycle;}

\title{\papertitle}

\title{Optical Music Recognition: State of the Art and Major Challenges}
\def\papertitle{Optical Music Recognition: State of the Art and Major Challenges }
\def\firstauthor{Elona Shatri}
\def\secondauthor{Gy\"orgy Fazekas}

 \twoauthors
 {\firstauthor} {Queen Mary University of London \\ %
 {\tt \href{mailto:e.shatri@qmul.ac.uk}{e.shatri@qmul.ac.uk}}}
 {\secondauthor} {Queen Mary University of London \\ %
 {\tt \href{mailto:g.fazekas@qmul.ac.uk}{g.fazekas@qmul.ac.uk}}}

\begin{document}\sloppy

\capstartfalse
\maketitle
\capstarttrue
\begin{abstract}

Optical Music Recognition (OMR) is concerned with transcribing sheet music into a machine-readable format. The transcribed copy should allow musicians to compose, play and edit music by taking a picture of a music sheet. Complete transcription of sheet music would also enable more efficient archival. OMR facilitates examining sheet music statistically or searching for patterns of notations, thus helping use cases in digital musicology too. Recently, there has been a shift in OMR from using conventional computer vision techniques towards a deep learning approach. In this paper, we review relevant works in OMR, including fundamental methods and significant outcomes, and highlight different stages of the OMR pipeline. These stages often lack standard input and output representation and standardised evaluation. Therefore, comparing different approaches and evaluating the impact of different processing methods can become rather complex. This paper provides recommendations for future work, addressing some of the highlighted issues and represents a position in furthering this important field of research.

\end{abstract}
\section{Introduction}\label{sec:introduction}

Music is often described as structured notes in time. Musical notations are systems that visually communicate this definition of music. The earliest known scores date back to 1250-1200 BC in Babylonia \cite{west1994babylonian}. Since then, many notation systems have emerged in different eras and different locations. Common Western Music Notation (CWMN) has become one of the most frequently used systems. This notation has evolved from the mensural music notation used before the seventeenth century. Current work in Optical Music Recognition focuses on the CWMN; nonetheless, studies are also carried out for old notations, including mensural, as shown in Table \ref{table:music-notation}.

\begin{table*}
    \centering
    \begin{tabular}{|p{9.1cm}|p{1.2cm}|p{0.8cm}|p{1.2cm}|p{1.7cm}|}
    \hline 
    References & CWMN & Old & Typeset & Handwritten \\
    \hline
    \hline
    Fujinaga \cite{fujinaga1988optical}, 
    Co{\"u}asnon et al. \cite{couasnon1995using}, 
    Ng and Boyle \cite{ng1996recognition}, 
    Chen et al. \cite{chen2010math},
    Vidal \cite{vidal2012staffline},
    Bui et al. \cite{bui2014staff}, 
    Huang et al. \cite{huang2019state} & \checkmark & & \checkmark & \\
    \hline 
    \hline
    Ng et al. \cite{ng1999embracing}, 
    Bainbridge and Bell \cite{Bainbridge2003AMN}, 
    Gocke \cite{gocke_building_nodate}, 
    Rebelo et al. \cite{shortest-path}, 
    Forn{\'e}s et al.\cite{fornes_use_2009}, 
    Pinto et al. \cite{pinto2011music}, 
    Haji{\v{c}} and Pecina \cite{hajivc2017muscima++}, 
    Roy et al. \cite{roy2017hmm}, 
    Pacha et al. \cite{pacha_handwritten_2018}, 
    Tuggener et al. \cite{watershed2018deep}, 
    Bar{\'o} et al. \cite{baro2019optical, baro2016towards} & 
    \checkmark & & & \checkmark \\
    \hline
    Calvo-Zaragoza and Rizo \cite{calvo2018end}, 
    Wen et al. \cite{wen2015new}, 
    Pacha and Eidenberger \cite{pacha_self, pacha2017towards}, 
    Calvo-Zaragoza et al. \cite{calvo2018camera} & \checkmark & & \checkmark & \checkmark \\
    \hline
    Calvo-Zaragoza et al. \cite{calvo2015staff-avoiding, mensural-calvo2017handwritten}, 
    Huang et al. \cite{huang2015automatic}, 
    Tard{\'o}n et al. \cite{tardon2009optical} & & \checkmark & & \checkmark \\
    \hline
    \hline
    \end{tabular}
    \footnotesize
    \caption{Studies conducted in CWMN (Common Western Music Notation), and old notations (used before the CWMN, mostly mensural notations}
    \label{table:music-notation}
\end{table*}

\begin{figure*}[ht]
    \centerline{
    \includegraphics[width=11cm]{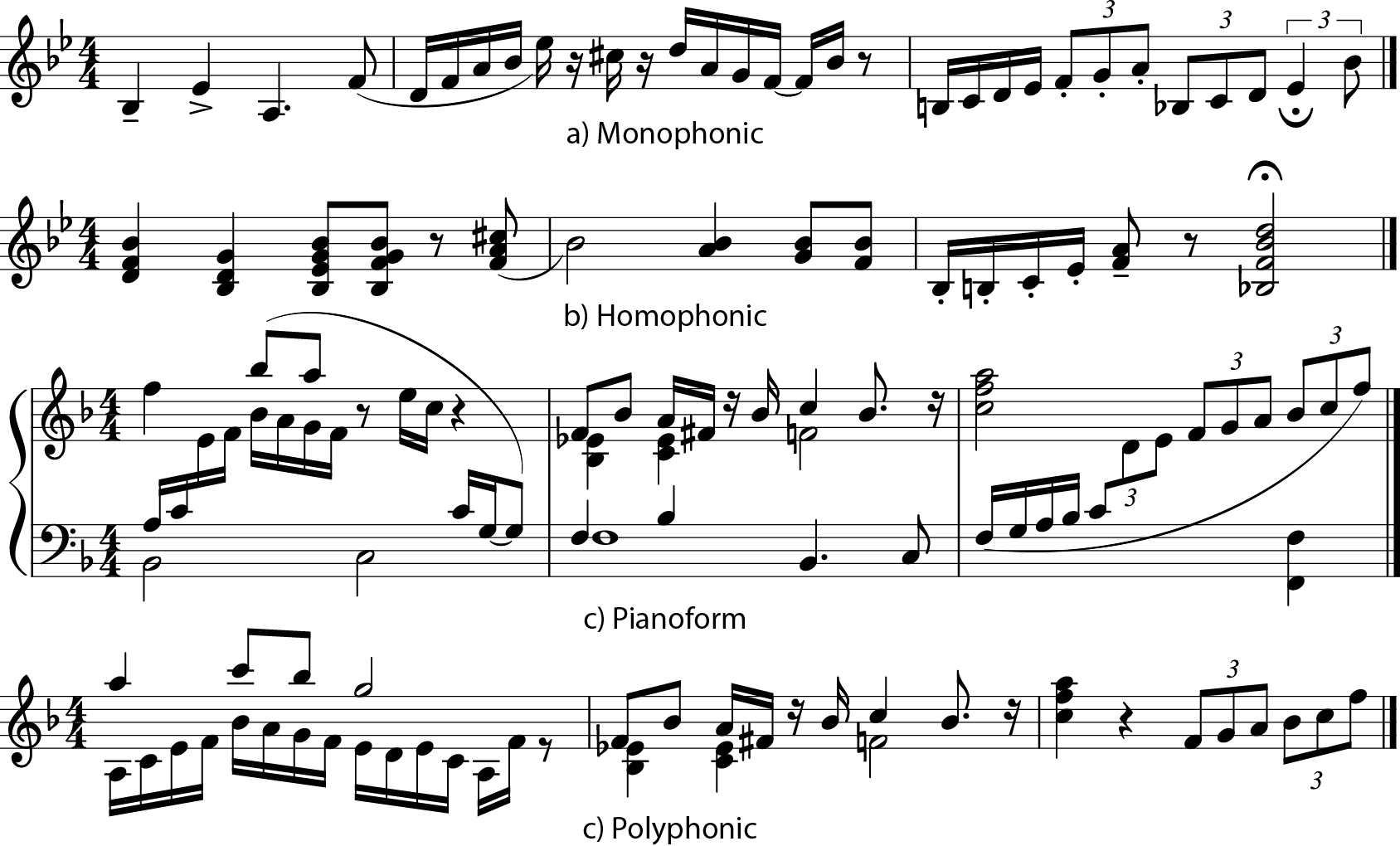}}
    \caption{A visual representation of the four categories of music notations \cite{byrd2015towards, calvo-zaragoza_understanding_2019}}
    \label{fig:monophonic}
\end{figure*}

Classifying music based on its difficulty is highly subjective. Nevertheless, Byrd and Simonsen \cite{byrd2015towards} in their attempt to have a standardised test-bed for OMR, name four categories based on the complexity of the score \cite{byrd2015towards} (see Figure \ref{fig:monophonic}):
\begin{enumerate}
 \item Monophonic: music in one staff with one note at a time;
 \item Polyphonic: multiple voices in one staff;
 \item Homophonic: multiple notes can occur at the same time to build up a chord, but only as a single voice;
 \item Pianoform: music in multiple staffs and multiple voices with significant structural interactions. 
\end{enumerate}
 
OMR has been researched for the last five decades; nonetheless, a unified definition of the problem is yet to emerge. However, Calvo-Zaragoza \cite{calvo-zaragoza_understanding_2019} offers the following definition of OMR.

\newtheorem{defn}{Definition}[section]
\begin{defn}
``Optical Music Recognition is a field of research that investigates how to computationally read music notation in documents.''
\end{defn}

The importance of OMR is evident both in the abundance of sheet music in archives and libraries, much of this is yet to be digitised, and in the common practice of musicians. Paper remains the first medium authors use to write music. By taking a picture of a score, OMR would enable us to later modify, play, add missing voices and share music using ubiquitous digital technologies. It also enables search capabilities, which are especially crucial for long pieces or large catalogues in music information retrieval and digital musicology. Other advantages of OMR include conversions to different sheet music formats (e. g. Braille music notation) and the ability to archive musical heritage \cite{jones2008optical}. 

Fundamentally, OMR's goal is to interpret musical symbols from images of sheet music. The output would be a transcribed version of the sheet, which is also machine-readable, i.e., musical symbols can be interpreted and manipulated computationally. The usual output formats are MusicXML and MIDI. These formats will include musical attributes and information such as pitches, duration, dynamics and notes. 

OMR has previously been referred to as Optical Character Recognition (OCR) for music. However, music scores carry information in a more complex structure, with ordered sequence of musical symbols together with their spatial relationships. In contrast, OCR deals with sequences of characters and words that are one-dimensional.  

Recently, the success deep learning has had in improving text and speech recognition has triggered a paradigm shift in OMR as well. One of the most comprehensive reviews on OMR was written in 2012 by Rebelo et al. \cite{rebelo_optical_2012}. However, at that time, the field had not yet seen the emergence of deep learning approaches. This position paper aims to update on these approaches.

State of the art works in OMR perform well with digitally written monophonic music, but there is plenty of room for improvement when it comes to reading handwritten music and complex pianoform scores \cite{calvo2018end, wen2015new, pacha_self}. The difficulty thus increases with the complexity of the music notation.

\begin{figure*}[ht]
 \centerline{
 \includegraphics[width=13cm]{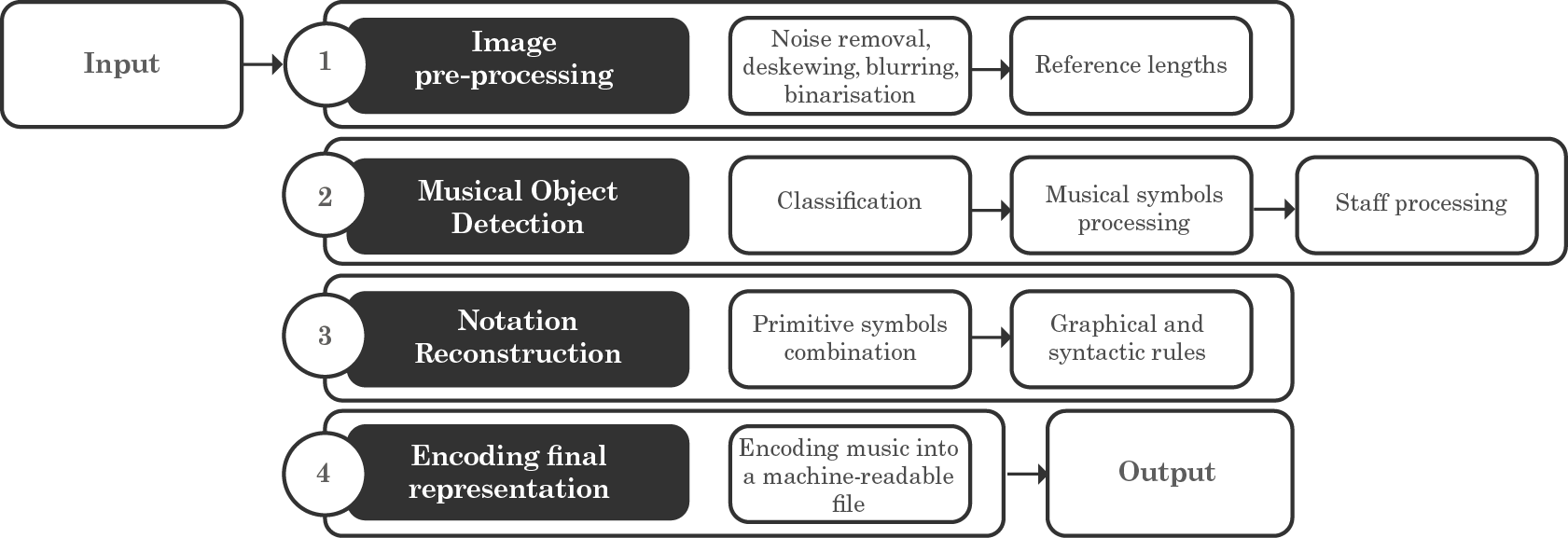}}
 \caption{Conventional OMR pipeline}
 \label{fig:OMR-framework}
\end{figure*}

\section{OMR Pipeline}

The standard OMR pipeline given by Rebelo et al. \cite{rebelo_optical_2012} is depicted in Figure \ref{fig:OMR-framework}:
\begin{enumerate}
 \item Image preprocessing;
 \item Music symbol recognition;
 \item Musical information reconstruction;
 \item Construction of a musical notation model.
\end{enumerate}
In the first stage, images of sheet music are subject to techniques such as noise removal, binarisation, de-skewing and blurring in order to make the rest of the OMR processes more robust. Subsequently, reference lengths, such as staff lines thickness and distances between them are calculated. Typically, the next stage is musical symbol recognition. This stage consists of staff line processing and musical symbol processing and ends with classification. Primitives of musical symbols  will be used in the third stage in order to reconstruct semantic meaning. Finally, all retrieved information should be embedded in an appropriate output file. A summary of these stages and the particular image processing and machine learning techniques employed in each stage are summarised in Table \ref{summary-of-studies}.

\begin{table*}
    \centering
    \begin{tabular}{|p{2.6cm}|p{13.5cm}|}
    \hline 
    Stage & Related Work \\
    \hline
    \hline
    Image preprocessing & 
    Fujinga \cite{fujinaga1988optical, fujinaga2004staff}, 
    Ng and Boyle \cite{ng1996recognition}, 
    Forn{\'e}s et al \cite{fornes2008writer, fornes_use_2009}, 
    Tard{\'o}n et al. \cite{tardon2009optical}, 
    Pinto et al. \cite{pinto2011music}, 
    Calvo-Zaragoza et al. \cite{calvo2015staff-avoiding}, 
    Huang et al. \cite{huang2015automatic}, 
    Wen et al. \cite{wen2015new}, 
    Ridler et al. \cite{ridler1978picture}, 
    Gocke \cite{gocke_building_nodate}, 
    Ballard \cite{ballard1981generalizing}, 
    Bainbridge and Bell \cite{Bainbridge2006}, 
    Cardoso et al. \cite{dos_santos_cardoso_staff_2009}, 
    Dalitz et al. \cite{dalitz_comparative_2008} \\
    \hline

    Symbol Recognition & 
    Mahoney \cite{mahoney1982automatic}, 
    Prerau \cite{prerau-grammar},
    Tard{\'o}n et al. \cite{tardon2009optical},
    Pacha \cite{incremental-apacha}, 
    Rebelo et al. \cite{shortest-path}, 
    Ng and Boyle \cite{ng1996recognition}, 
    Choudhury et al. \cite{choudhury2000optical},
    Bainbridge and Bell \cite{Bainbridge2003AMN}, 
    Forn{\'e}s et al. \cite{fornes2008writer, fornes_use_2009}, 
    Huang et al. \cite{huang2015automatic}
    Fujinaga \cite{fujinaga1988optical}, 
    Wen et al. \cite{wen2015new},
    Pacha et al. \cite{pacha_handwritten_2018, pacha2017towards},
    Chen et al. \cite{chen2010math}, 
    Gocke \cite{gocke_building_nodate}, 
    Miyao and Nakano \cite{miyao1996note} \\
    \hline

    Musical Information Reconstruction &
    Prerau \cite{prerau-grammar},
    Pacha et al. \cite{pacha_learning_2019, mensural_pacha},
    Roy et al. \cite{roy2017hmm}, 
    Bainbridge and Bell \cite{Bainbridge2003AMN}, 
    Co{\"u}asnon et al. \cite{couasnon1995using},
    Ng and Boyle \cite{ng1996recognition},
    Bar{\'o} et al. \cite{baro2017optical, baro2019optical}
    Calvo-Zaragoza et al. \cite{calvo2018end, calvo2018camera, mensural-calvo2017handwritten} \\
    \hline
    
    Musical Notation Model & 
    Droettboom et al. \cite{droettboom2002optical},
    Chen et al. \cite{chen2010math},
    Choudhury et al. \cite{choudhury2000optical}, Ng et al. \cite{ng1999embracing}, 
    Tard{\'o}n \cite{tardon2009optical}, Bainbridge and Bell \cite{Bainbridge2003AMN},
    Huang et al. \cite{huang2015automatic} \\

    \hline
    \hline
    \end{tabular}
    \footnotesize
    \caption{Summary of the studies carried in each of the OMR pipeline stages}
    \label{summary-of-studies}
\end{table*}

\section{Image preprocessing}

Image preprocessing is a fundamental step in many computer vision tasks. The primary outcome of this stage is an adjusted image that is easier to manipulate. Most common image manipulations include enhancement, de-skewing, blurring, noise removal and binarisation \cite{fujinaga1988optical, ng1996recognition, fornes2008writer, fornes_use_2009, tardon2009optical, pinto2011music, calvo2015staff-avoiding, huang2015automatic, wen2015new}. Image enhancement can include filters and adjusting the contrast or brightness for optimal object detection. De-skewing eliminates skewness and helps in obtaining a more appropriate view in the object detection stage. Most of the digital images during the acquisition, transmission or processing are subject to noise. Both colour and brightness contain signals that carry random noise. Depending on the features of the image, different types of filters are used to remove some of the noise. 
During the process of binarisation, images are analysed to decide what is noise and what constitutes useful information for the task. Techniques to choose a binarisation threshold include global and adaptive methods. A global threshold is typically determined for the whole image, while for the adaptive threshold, local information in the image should be considered. Ng et al. \cite{ng1996recognition} adapt the global threshold proposed by Ridler and Calvard \cite{ridler1978picture}. While adaptive threshold is used in several more recent OMR studies too \cite{fornes_use_2009, fornes2008writer, gocke_building_nodate}.

Gocke's \cite{gocke_building_nodate} pipeline starts with a Gaussian filter, thenceforth a histogram on each colour channel is built. Then, the image is rotated to find the best angle that maximises horizontal projections. The image is then segmented into smaller 30x30 pixels, and a local threshold is found for each tile. Following threshold selection, all elements smaller than 4 pixels in diameter are removed, making the image clearer. The image is finally ready for staff-removal and symbol recognition. Local thresholding in this case yielded better results than the global one. 

Similarly, in Forn{\'e}s et al. \cite{fornes_use_2009} binarisation is followed by de-skewing using the Hough Transform \cite{ballard1981generalizing}. A coarse approximation of the staff lines is obtained using median filters with horizontal masks to reconstruct the staff lines later. However, in this process, some residual colour information is retained, especially where the lines intersect with musical symbols, hence, some noise is still left.  This approach is not robust to damaged paper. 
    
Pinto et al. \cite{pinto2011music} propose a content-aware binarisation method for music scores. The model captures content-related information during the process from a greyscale image. It also extracts the staff line thickness and the vertical line distance in staff to guide binarisation. This algorithm tries to find a threshold that maximises the extracted content information from images. However, the performance hugely depends on the document characteristics, limiting performance across different documents.

Calvo-Zaragoza and Gallego \cite{calvo2019selectional, gallego2017staff} propose using selectional auto-encoders \cite{masci2011stacked} to learn an end-to-end transformation for binarisation. The network activation nodes indicate the likelihood of whether pixels are foreground or background pixels. Ensuing training, documents are parsed through the model and binarised using an appropriate global threshold. This approach performs better than the conventional binarisation methods in some document types. Nonetheless, errors happen around foreground strokes and are emphasised along edges of the input windows, due to the lack of context in the neighbourhood.

\section{Music Symbol Recognition}

The next stage typically constitutes dealing with musical symbol recognition. Here, the three main steps are staff processing, isolating musical symbols and finally, classification. Usually, staff lines are first detected and then removed from the images. The model then isolates the remaining notations as primitive elements. These are later used to extract features and feed those features to train the classifier. 

\subsection{Staff processing}

Staff lines are a set of five horizontal lines from one side of the music score to the other. Each line and gap represent a different pitch. For better object detection, the question of staff line removal has been of prime importance. Researchers take two different approaches; one is only detecting and isolating them, while the other approach goes one step further in removing them. 

While in printed sheet music, staff lines are straight, parallel and horizontal, in handwritten scores, these lines might be tilted, curved and may not be parallel at all. These lines might also look curved or skewed depending on the image skew angle \cite{shortest-path} or the degradation of the paper. The model needs to separate staff lines from actual music objects. Since the lines overlap with musical objects, simply cutting and removing them degrades the notes and make them harder to recognise, further limiting performance. 

Consequently, an increasing number of studies take the approach of removing the staff lines in a more intelligent fashion \cite{ng1996recognition, choudhury2000optical, blostein1992critical,Bainbridge2003AMN, fornes2008writer, fornes_use_2009, tardon2009optical, huang2015automatic, wen2015new}. In this section, we outline typical staff line processing approaches. Blostein and Baird \cite{blostein1992critical} suggests using horizontal projections of the black pixels and finding their maxima. The drawback is that the method only considers horizontal straight lines. In order to deal with non-horizontal, the process is followed with image rotations and choosing an angle with a higher maxima.

Rebelo et al. (2007) \cite{shortest-path} consider staff lines to be the shortest path between two horizontal page margins if those paths have black pixels throughout the entire path. The height between every two lines is first estimated and later used as a reference length for the following operations. Upon choosing an estimation, using the Dijkstra algorithm \cite{dijkstra1959note}, the shortest path between the leftmost pixel and the rightmost pixel is found. Their method is robust to lines with some curvature and discontinuity since it follows continuous paths connecting line ends from both sides. However, this algorithm may sometimes retain paths that do not follow the staff line. This happens when there is a higher density of beamed notes, and the estimated path follows the beams or when the staff lines are very curved. 

Cardoso et al. \cite{dos_santos_cardoso_staff_2009} propose stable paths, considering the sheet music image as a graph. The staff lines in the graph are the less costly paths between the left and right margins. Subsequently, the model should differentiate between score pixels and staff line pixels. This model is robust to discontinuities, skewness, curvature in staff lines and one-pixel thin staff lines. Both the shortest path and stable paths give a similar false detection rate in test set of 32 ideal score images. This set is subject to different deformations, resulting in 2,688 total images. However, the stable path approach is five times faster. This technique is often used in the preprocessing stage \cite{mensural-calvo2017handwritten}.

Another study \cite{bui2014staff} uses stable paths approach to extract staff line skeletons. Then, the line adjacency graph (LAG) \cite{iliescu1996proposed} is used to cluster pixel runs generated from run-length encoding (RLE) of the image \cite{tsukiyama1986method}. The last step involves removing clusters lying on the staff line. This step has two passes; the first step estimates the height line for each staff by averaging the section height being cut with the staff lines. The second pass filters out the noise left from the last pass. This method takes a similar approach with \cite{carter1992automatic} grouping staff line pixels into segments.

Other studies follow the approach of keeping the staff lines during the next stages \cite{ng1999embracing, gocke_building_nodate, pugin2006optical, calvo2015staff-avoiding, roy2017hmm, calvo2016early}. They argue that the staff line removal task is very complex and often ends up being inaccurate and passes errors to the following stages. These studies usually detect and isolate staff lines ahead of object processing. Recent object detection studies show that removing staff lines does not add much improvement to this stage \cite{pacha_handwritten_2018}.

 \begin{figure*}[ht]
 \centerline{
 \includegraphics[width=13cm, ]{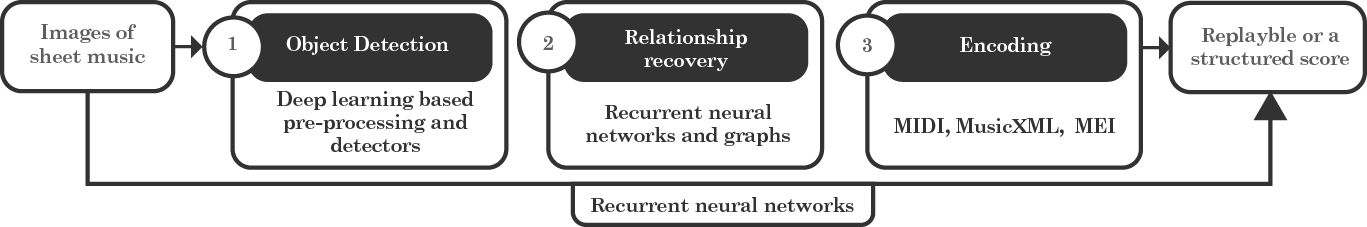}}
 \caption{Typical OMR pipeline using deep neural networks}
 \label{fig:dl-learning-framework}
\end{figure*}
A more recent work \cite{incremental-apacha} investigates how incremental learning can assist staff line detection using convolutional neural networks (CNNs) and human annotation. To begin with, a CNN model is fed a small amount of data with available annotations for training. Using this training, the model makes predictions on a larger dataset, and a human annotator rejects or accepts the predictions. The accepted predictions are added to the training dataset to repeat the process. This method enables the creation of a more extensive dataset. After four iterations, the dataset contains 70\% annotated scores of the original set. One drawback of incremental learning is that if the annotator accepts samples with imperfect annotations, the error accumulates in each iteration, introducing inaccuracy, while it also needs a human annotator. This yields similar results with \cite{dos_santos_cardoso_staff_2009, carter1992automatic}, however, different evaluation metrics are used. 

Despite the substantial research effort put into staff line removal, it is still far from being accurate in handwritten sheet music. Handwritten scores exhibit a wide variety in line length and distance, thickness, curvatures of staff lines and also the quality of the image.

\subsection{Music symbol processing}

The next step after removing the staff lines is to isolate the musical symbols. Staff line removal will strongly affect this step as it can cause fragmentation in the parts where staff lines and musical objects are tangent to each other. One widely used approach is hierarchical decomposition \cite{rebelo_optical_2012}, where staff lines split a music sheet and then extract noteheads, rests stems and other notation elements \cite{choudhury2000optical, droettboom2002optical, gocke_building_nodate, miyao1996note, ng1996recognition}. Some approaches consider, for instance, a half-note instead of its primitives for the classification step. Mahoney \cite{mahoney1982automatic} uses descriptors to choose the matching candidate between a set of candidates of symbol types. Carter \cite{carter1989automatic} uses the line-adjacency graph (LAG) of an image for both removing the staff lines and providing a structural analysis of symbols. This technique helps in obtaining more consistent image sectioning, but it is limited to a small range of symbols as well as a potentially severe break-up of symbols. 

Some studies skip segmentation and staff line removal \cite{pugin2006optical, calvo2016early, roy2017hmm} and use Hidden Markov Models (HMM). HMMs work on low-level features that are robust to poor quality images and can detect early topographic prints and handwritten pieces. Calvo-Zaragoza \cite{calvo2016early} split sheet music pages into staves following preprocessing. All staves are normalised and later represented as a sequence of feature vectors. This approach is very similar to \cite{pugin2006optical}, however, this study goes one step further and supports the HMM with a statistical N-gram model and achieve a 30\% error rate. This performance could be further improved if lyrics are removed, light equalisation is performed and data variations are statistically modelled.

\subsection{Music symbol classification}

After the segmentation of musical primitives, the subsequent process is classification. Objects are classified based on their shapes and similarities. However, since these objects are very often densely packed and overlapping their shapes can become very complex. Therefore, this step is very sensitive to all possible variations in music notations. 
Fujinaga \cite{fujinaga1988optical} uses projection profiles for classification, Gocke \cite{gocke_building_nodate} uses template matching to classify the objects. Other methods used are support vector machines (SVMs), k-nearest neighbour (kNN), neural networks (NN) and hidden Markov models (HMM). A comparative study of the four methods \cite{rebelo2010optical}, finds SVM performs better than HMMs.

Considering the success of deep neural networks (DNN) in many machine learning tasks, recent studies take this approach in music object recognition and classification. A typical pipeline is shown in Figure~\ref{fig:dl-learning-framework}. These networks have many layers with activation functions employed before information propagates to the next layer. The deeper the model, the more complicated it gets and is able to detect hidden nonlinear relationships between the data, in this case, music objects. The problem with using DNNs in OMR is that they require a significant amount of labelled data for supervised training.

Object detection in images is a very active research field. Regional CNNs (R-CNNs), Faster R-CNN \cite{ren2015faster}, U-nets \cite{hajic2018towards}, deep watershed detectors \cite{watershed2018deep} and Single-shot detectors \cite{dai2016r, liu2016ssd} are among some of the approaches proposed recently. Pacha et al. \cite{pacha_handwritten_2018} use Faster R-CNN networks with pre-trained models fine-tuned with data from MUSCIMA++ (see Sect.~\ref{sec:datasets} for a summary of OMR datasets). They achieve a mean average precision of up to 80 \%. However, such performance is achieved with cropping the image into individual staff lines.

Tuggener et al. \cite{watershed2018deep} use deep watershed detectors in the whole image. It is faster than Faster R-CNN approach in image snippets, and it allows some shift in the data distribution. Nonetheless, it does not perform well on underrepresented classes.

Going further into the pipeline, we should be able to capture and reconstruct the right positions, relationships between notes, and relevant musical semantic information such as duration, onsets, pitch.

\section{Notation Reconstruction}

After classifying and recognising musical objects, the next block should extract musical semantics and structure. As mentioned earlier, OMR is two-dimensional, meaning that recognising the note sequence as well as their spatial relationships are essential. Hence, a model should identify the information about the spatial relationship between the recognised objects.
Ng et al. \cite{ng1999embracing} believe that domain knowledge is key to improving OMR tasks and especially music object recognition, similarly to a trained copyist or engraver, to decipher poorly written scores, building on the authors' previous research on printed scores \cite{ng1996recognition}. A multi-stage process is adopted, in which the first search is for essential features helping the interpretation of the score, verified by their mutual coherence, followed by a more intelligent search for more ambiguous features. Key and time signatures are detected after low-level processing and classification, using these global high-level features to test the earlier results. 

Ng and Boyle \cite{ng1996recognition} base their study on three assumptions: i) foreknowing the time signature, ii) key signature, and iii) that the set of the primitive feature set under examination is limited to ten. The first and second assumptions are overcome by geometrically predicting a limited symbol set such as numbers, flats and sharps. The input image goes through binarisation using a threshold, image rotation for de-skew, then the staff lines are detected and erased. Now the image has blocks of pixels, music object primitives and groups of primitives. Further segmentation based on some rules is needed for a group of primitives. After the segmentation process, a classifier uses only the width and the height of the bounding box for recognition based on a sampled training set. The recognised primitives are grouped to reconstruct their semantic meaning. The reconstruction consists of overlaying an ellipse and counting the number of foreground pixels, finding the pitch, search the neighbourhood for other features that might belong to the object and identifying the possible accidents using a nearest neighbourhood (NN) classifier. Music knowledge related to bars, time, and key signatures is applied at this stage. During segmentation, the process relies on straight edges of the objects, therefore is not robust to handwritten scores. The method fails if the symbols are skewed, for instance, when a stem is not perpendicular to a stave line. 

Similar to the method mentioned above, another approach is formalising musical knowledge and/or encoding knowledge into grammar rules that explain, for instance, how primitives are to be processed or how graphical shapes are to be segmented \cite{Bainbridge2003AMN, couasnon1995using}.

Prerau\cite{prerau-grammar} proposes two levels of grammar. One being notational grammar while the other is a higher-level grammar for music. The first allows the recognition of symbol relationships, the second deals with larger music units.
Many other techniques use musical rules to create grammar rules for OMR. Such rules can be exemplified as \cite{rebelo_optical_2012}:
\begin{itemize}
\item An accidental is placed before a notehead and at the same height;
\item A dot is placed after or above a notehead in a variable distance;
\item Between any pair of symbols: they cannot overlap.
\end{itemize}

The issue with music rules and heuristics is that these rules are very often violated, especially in handwritten music. Furthermore, it is challenging to create rules for many different variations and notations with a high level of complexity. As a result, this approach would not perform well with both typeset and handwritten complex notations, and it is difficult to scale to a broad range of notation and engraving styles. 

Pacha et al. \cite{pacha_learning_2019} propose using graphs to move towards a universal music representation. Considering that in music notations, the relationship between primitives contains the semantic meaning of each primitive; they suggest that OMR should employ a notation assembly stage to represent this relationship. Instead of using grammar and rules mentioned earlier, they use a machine learning approach to assemble a set of detected primitives. The assembly is similar to a graph containing syntactic relationships among primitives capturing the symbol configuration. The robustness of the model regarding variations in bounding boxes leaves room for improvement and so does the notation assembly stage, due to the lack of broader hypotheses on the detected objects.  

Bar{\'o} et al. \cite{baro2017optical} consider monophonic scores as sequences and use Long Short-Term Memory (LSTM) Recurrent Neural Networks (RNNs) for reading such sequences to retrieve pitch and duration. For evaluation they use Symbol Error Rate (SER) defined as the minimum number of edit operation to convert an array to another. This approach shows to work well with simple scores such as monophonic scores, but fundamental remodelling is needed for more complex scores \cite{baro2017optical}.

This stage is concerned with reconstructing relationships from the detected musical objects. A challenge in this stage is to model a musical output representation that encodes sheet music both a similar rendering of the original image and the semantics (e.g. onsets, duration, pitch).

\vspace{-5pt}
\section{Music Notation Encoding}

The output from the previous steps is used to construct a semantic model or data model. This model should represent a re-encoding of the score in the input. The output model should be expressible in a machine-readable format. Usual OMR output formats include MIDI, MusicXML, MEI, NIFF, Finale, and in some software, the music is even rendered into WAVE files. Musical Instrument Digital Interface (MIDI) \cite{rothstein1995midi} is an interchange medium between the computer and digital instruments. At the basic level, MIDI includes the temporal position when a note starts, stops, how loud the note is, the pitch of the note, instrument and channel. The main drawback of MIDI is that it cannot represent the relationships between musical symbols, or produce a re-encoded structured file, limiting the output to replayability only.

Notable formats that allow a structured encoding and storing notations include MusicXML \cite{good2001musicxml,good2006lessons} and MEI \cite{roland2002music, hankinson2011music}. Both allow further editing in a music notation software. MusicXML is more focused on encoding notation layout. It is designed for archiving and for sharing sheet music between applications. There is ongoing research in the W3C Music Notation Community Group on improving MusicXML format to handle more specific tasks and applications. 

The Music Encoding Initiative (MEI) \cite{roland2002music} claims to be comprehensive, declarative, explicit and hierarchical. MEI has not been widely used as the final output of OMR systems yet. However, based on the characteristics mentioned above, MEI is able to capture and retain musical semantics better, e.g. relationships between voices, which may benefit music engraving. 

There is also work converting OMR output into Semantic Web formats. Jones et. al.~\cite{jones2017musicowl} propose the use of Linked Data to annotate and improve discovery of music scores using the Resource Description Framework (RDF). The captured information is limited to the number of voices, movements and melodies. Further extensions are needed to store more sophisticated music semantics that support harmony or melody analysis. Nevertheless, the use of Linked Data compatible formats may benefit OMR applications in multiple ways. Linking scores to other music related data on the Web~\cite{fazekas2010an} or even features of the audio of a performance~\cite{allik2016ontology} could support interactive applications such as score following or large catalogue navigation~\cite{turchet2019cloud}. The ontologies governing these formats may be used to encode musical or engraving rules to complement probabilistic inference in machine learning models. 

To decide which of the encodings to use, we have to think of what an application may require. Using the knowledge obtained in the previous steps and from different studies would assist this stage in its standardisation. Currently there is little research in OMR dealing with encoding, however, many works in other fields focus on encoding formats that better represent music and its structure. 

\vspace{-10pt}
\section{Datasets}
\label{sec:datasets}
Depending on the OMR task to be performed and the nature of the application, different datasets may be suitable. Existing datasets contain handwritten or copyright-free printed music sheets in mensural or CWMN notations. 
Calvo-Zaragoza et al. \cite{calvo2014recognition} introduced a new dataset called HOMUS (Handwritten Online Musical Symbols). This contains 15200 samples of 32 types of musical symbols from 100 different musicians. 
Universal Music Symbol Collection is a dataset of 90000 tiny handwritten and typeset music symbols from 79 classes that can be used to train classifiers. 

As for staff line removal, a commonly used dataset is CVC-MUSCIMA \cite{fornes2012cvc}. It contains 1000 music sheets written by 50 different musicians. Each musician was asked to transcribe the same given 20 pages of music using the same pen and same style of sheet music paper. These pages include monophonic and polyphonic music, consisting of scores for solo instruments and music scores for choir and orchestra.

A derived version of CVC-MUSCIMA dataset is MUSCIMA++ \cite{hajivc2017muscima++}. This dataset is more suitable for musical symbol detection. It has 91255 symbols with both notation primitives and higher-level notation objects, key signatures or time signatures. Notes are captured using the annotated relationships of the primitives, having this way both low and high-level symbols.
DeepScores is a collection that contains 300k annotated images of written music mainly for object classification, detection, and segmentation \cite{tuggener2018deepscores}. This dataset has large images containing tiny objects. 

There are also datasets for an end-to-end recognition such as the Printed Images of Music Staves (PrIMuS) \cite{calvo2018end}, or the extended version of this with distorted images to simulate imperfections Camera-PrIMuS \cite{calvo2018camera}. These datasets have 87678 real-music scripts in five different formats: PNG, MIDI, MEI, semantic and agnostic encoding which is a sequence that contains the graphical symbols and their positions without any musical meaning. 

Given that the performance of the deep learning methods usually depends on the amount of the data the model is fed, for future work, we propose creating a universal dataset that facilitates the intermediate stages but also an end-to-end system. We want to start by generating music files using a music notation software such as Dorico \cite{dorico} or Rosegarden \cite{rosegarden}. This work will be harmonized with the before-mentioned MUSCIMA++ and DeepScores datasets.

\vspace{-10pt}
\section{Open Issues and Conclusions}

Low-quality images of sheet music, complex scores, handwritten music and alternate notations are still challenging for OMR, while most of the work focuses on monophonic scores. CWMN notation is highly complex, having dense scores, overlapping symbols, structural complexity, semantic rules that are sometimes violated. For a deep learning approach, in particular, class imbalance is one of the most significant issues; some note types are persistent while some others are rare. An further open issue is the lack of a large labelled dataset with a broad variety of image quality and balanced classes \cite{novotny2015introduction}. 

We can observe a shift in OMR from using conventional image processing and object detection to using neural networks, as shown in Figure \ref{fig:dl-learning-framework}. Recently published papers take novel approaches and use deep learning methods in all stages of the OMR pipeline. These stages are not necessarily in the order presented above or exhibit all the steps described.

Despite the introduction of deep learning, the field leaves space for improvement in all stages of the pipeline. New opportunities include creating more diverse and better balanced datasets, improving the detection of music objects and staff lines, the reconstruction of semantic meaning, and, perhaps most importantly, standardising the evaluation metrics and the output of the pipeline. A possible final goal is end-to-end learning that would not need intermediate steps. Neural networks are already applied to problems like text and speech recognition and machine translation in this manner. However, these systems are still not adapted to a two-dimensional output sequence such as music \cite{calvo-zaragoza_understanding_2019}.

This paper summarised seminal and influential studies conducted in the field of OMR. We discussed different methods and approaches in prominent stages of the OMR pipeline. Our review aims to identify important older works and current state-of-the-art approaches, which can be used as a reference by researchers to begin further work in OMR. It also represents a position in several aspects of the field, including the need for incorporating more prior knowledge, theory and musical information in the processing pipeline, the need for finding new methods to incorporate these priors into statistical learning models such as deep neural networks and a need for more standardisation in OMR evaluation.

\begin{acknowledgments}
 The authors acknowledge the support of the AI and Music CDT, funded by UKRI and EPSRC under grant agreement no. EP/S022694/1 and our industry partner Steinberg Media Technologies GmbH for their continuous support. 
\end{acknowledgments}


\bibliography{tenor2020-template}

\begin{thebibliography}{10}
\providecommand{\url}[1]{#1}
\csname url@samestyle\endcsname
\providecommand{\newblock}{\relax}
\providecommand{\bibinfo}[2]{#2}
\providecommand{\BIBentrySTDinterwordspacing}{\spaceskip=0pt\relax}
\providecommand{\BIBentryALTinterwordstretchfactor}{4}
\providecommand{\BIBentryALTinterwordspacing}{\spaceskip=\fontdimen2\font plus
\BIBentryALTinterwordstretchfactor\fontdimen3\font minus
  \fontdimen4\font\relax}
\providecommand{\BIBforeignlanguage}[2]{{%
\expandafter\ifx\csname l@#1\endcsname\relax
\typeout{** WARNING: IEEEtran.bst: No hyphenation pattern has been}%
\typeout{** loaded for the language `#1'. Using the pattern for}%
\typeout{** the default language instead.}%
\else
\language=\csname l@#1\endcsname
\fi
#2}}
\providecommand{\BIBdecl}{\relax}
\BIBdecl

\bibitem{west1994babylonian}
M.~L. West, ``The babylonian musical notation and the hurrian melodic texts,''
  \emph{Music \& letters}, vol.~75, no.~2, pp. 161--179, 1994.

\bibitem{fujinaga1988optical}
I.~Fujinaga, ``Optical music recognition using projections,'' Ph.D.
  dissertation, McGill University, 1988.

\bibitem{couasnon1995using}
B.~Co{\"u}asnon, P.~Brisset, and I.~St{\'{e}}phan, ``Using logic programming
  languages for optical music recognition,'' in \emph{3rd International
  Conference on the Practical Application of Prolog}, 1995.

\bibitem{ng1996recognition}
K.~Ng and R.~Boyle, ``Recognition and reconstruction of primitives in music
  scores,'' \emph{Image and Vision Computing}, vol.~14, no.~1, pp. 39--46,
  1996.

\bibitem{chen2010math}
G.~{Chen}, L.~{Zhang}, W.~{Zhang}, and Q.~{Wang}, ``Detecting the staff-lines
  of musical score with hough transform and mathematical morphology,'' in
  \emph{2010 International Conference on Multimedia Technology}, Oct 2010, pp.
  1--4.

\bibitem{vidal2012staffline}
A.~Rebelo and J.~d.~S. Cardoso, ``Staff line detection and removal in the
  grayscale domain,'' Ph.D. dissertation, 2013.

\bibitem{bui2014staff}
H.-N. Bui, I.-S. Na, and S.-H. Kim, ``Staff line removal using line adjacency
  graph and staff line skeleton for camera-based printed music scores,'' in
  \emph{22nd International Conference on Pattern Recognition}, 2014, pp.
  2787--2789.

\bibitem{huang2019state}
Z.~Huang, X.~Jia, and Y.~Guo, ``State-of-the-art model for music object
  recognition with deep learning,'' \emph{Applied Sciences}, vol.~9, no.~13,
  pp. 2645--2665, 2019.

\bibitem{ng1999embracing}
K.~Ng, D.~Cooper, E.~Stefani, R.~Boyle, and N.~Bailey, ``Embracing the composer
  : Optical recognition of handwrtten manuscripts,'' in \emph{International
  Computer Music Conference}, 1999, pp. 500--503.

\bibitem{Bainbridge2003AMN}
D.~Bainbridge and T.~Bell, ``A music notation construction engine for optical
  music recognition,'' \emph{Software: Practice and Experience}, vol.~33,
  no.~2, pp. 173--200, 2003.

\bibitem{gocke_building_nodate}
R.~G{\"{o}}cke, ``Building a system for writer identification on handwritten
  music scores,'' pp. 250--255, 2003.

\bibitem{shortest-path}
A.~Rebelo, A.~Capela, J.~F. Pinto~da Costa, C.~Guedes, E.~Carrapatoso, and
  J.~d.~S. Cardoso, ``A shortest path approach for staff line detection,'' in
  \emph{3rd International Conference on Automated Production of Cross Media
  Content for Multi-Channel Distribution}, 2007, pp. 79--85.

\bibitem{fornes_use_2009}
A.~Forn{\'{e}}s, J.~Llad{\'{o}}s, G.~S{\'{a}}nchez, and H.~Bunke, ``On the use
  of textural features for writer identification in old handwritten music
  scores,'' 2009, pp. 996--1000.

\bibitem{pinto2011music}
T.~Pinto, A.~Rebelo, G.~Giraldi, and J.~d.~S. Cardoso, ``Music score
  binarization based on domain knowledge,'' in \emph{Pattern Recognition and
  Image Analysis}, J.~Vitri{\`a}, J.~M. Sanches, and M.~Hern{\'a}ndez,
  Eds.\hskip 1em plus 0.5em minus 0.4em\relax Springer Berlin Heidelberg, 2011,
  pp. 700--708.

\bibitem{hajivc2017muscima++}
J.~Haji{\v{c}}~jr. and P.~Pecina, ``The {MUSCIMA++} dataset for handwritten
  optical music recognition,'' in \emph{14th International Conference on
  Document Analysis and Recognition}, Kyoto, Japan, 2017, pp. 39--46.

\bibitem{roy2017hmm}
P.~P. Roy, A.~K. Bhunia, and U.~Pal, ``{HMM}-based writer identification in
  music score documents without staff-line removal,'' \emph{Expert Systems with
  Applications}, vol.~89, pp. 222--240, 2017.

\bibitem{pacha_handwritten_2018}
A.~Pacha, K.-Y. Choi, B.~Co{\"{u}}asnon, Y.~Ricquebourg, R.~Zanibbi, and
  H.~Eidenberger, ``Handwritten music object detection: Open issues and
  baseline results,'' in \emph{13th International Workshop on Document Analysis
  Systems}, 2018, pp. 163--168.

\bibitem{watershed2018deep}
L.~Tuggener, I.~Elezi, J.~Schmidhuber, and T.~Stadelmann, ``Deep watershed
  detector for music object recognition,'' in \emph{19th International Society
  for Music Information Retrieval Conference}, Paris, France, 2018, pp.
  271--278.

\bibitem{baro2019optical}
A.~Bar{\'{o}}, P.~Riba, J.~Calvo-Zaragoza, and A.~Forn{\'{e}}s, ``From optical
  music recognition to handwritten music recognition: A baseline,''
  \emph{Pattern Recognition Letters}, vol. 123, pp. 1--8, 2019.

\bibitem{baro2016towards}
A.~Bar{\'{o}}, P.~Riba, and A.~Forn{\'{e}}s, ``Towards the recognition of
  compound music notes in handwritten music scores,'' in \emph{15th
  International Conference on Frontiers in Handwriting Recognition}.\hskip 1em
  plus 0.5em minus 0.4em\relax Institute of Electrical and Electronics
  Engineers Inc., 2016, pp. 465--470.

\bibitem{calvo2018end}
J.~Calvo-Zaragoza and D.~Rizo, ``End-to-end neural optical music recognition of
  monophonic scores,'' \emph{Applied Sciences}, vol.~8, no.~4, 2018.

\bibitem{wen2015new}
C.~Wen, A.~Rebelo, J.~Zhang, and J.~d.~S. Cardoso, ``A new optical music
  recognition system based on combined neural network,'' \emph{Pattern
  Recognition Letters}, vol.~58, pp. 1--7, 2015.

\bibitem{pacha_self}
A.~Pacha and H.~Eidenberger, ``Towards self-learning optical music
  recognition,'' in \emph{16th International Conference on Machine Learning and
  Applications}, 2017, pp. 795--800.

\bibitem{pacha2017towards}
------, ``Towards a universal music symbol classifier,'' in \emph{14th
  International Conference on Document Analysis and Recognition}, IAPR TC10
  (Technical Committee on Graphics Recognition).\hskip 1em plus 0.5em minus
  0.4em\relax Kyoto, Japan: IEEE Computer Society, 2017, pp. 35--36.

\bibitem{calvo2018camera}
J.~Calvo-Zaragoza and D.~Rizo, ``Camera-primus: Neural end-to-end optical music
  recognition on realistic monophonic scores,'' in \emph{19th International
  Society for Music Information Retrieval Conference}, Paris, France, 2018, pp.
  248--255.

\bibitem{calvo2015staff-avoiding}
J.~Calvo-Zaragoza, I.~Barbancho, L.~J. Tard{\'o}n, and A.~M. Barbancho,
  ``Avoiding staff removal stage in optical music recognition:{~}application to
  scores written in white mensural notation,'' \emph{Pattern Analysis and
  Applications}, vol.~18, no.~4, pp. 933--943, 2015.

\bibitem{mensural-calvo2017handwritten}
J.~Calvo-Zaragoza, A.~Toselli, and E.~Vidal, ``Handwritten music recognition
  for mensural notation: Formulation, data and baseline results,'' in
  \emph{14th International Conference on Document Analysis and Recognition},
  Kyoto, Japan, 2017, pp. 1081--1086.

\bibitem{huang2015automatic}
Y.-H. Huang, X.~Chen, S.~Beck, D.~Burn, and L.~Van~Gool, ``Automatic
  handwritten mensural notation interpreter: From manuscript to {MIDI}
  performance,'' in \emph{16th International Society for Music Information
  Retrieval Conference}, M.~M{\"{u}}ller and F.~Wiering, Eds., M{\'{a}}laga,
  Spain, 2015, pp. 79--85.

\bibitem{tardon2009optical}
L.~J. Tard{\'o}n, S.~Sammartino, I.~Barbancho, V.~G{\'o}mez, and A.~Oliver,
  ``Optical music recognition for scores written in white mensural notation,''
  \emph{EURASIP Journal on Image and Video Processing}, vol. 2009, no.~1, p.
  843401, 2009.

\bibitem{byrd2015towards}
D.~Byrd and J.~G. Simonsen, ``Towards a standard testbed for optical music
  recognition: Definitions, metrics, and page images,'' \emph{Journal of New
  Music Research}, vol.~44, no.~3, pp. 169--195, 2015.

\bibitem{calvo-zaragoza_understanding_2019}
J.~Calvo-Zaragoza, J.~Haji{\v{c}}~jr., and A.~Pacha, ``Understanding optical
  music recognition,'' \emph{Computing Research Repository}, 2019.

\bibitem{jones2008optical}
G.~Jones, B.~Ong, I.~Bruno, and K.~Ng, ``Optical music imaging: Music document
  digitisation, recognition, evaluation, and restoration,'' in
  \emph{Interactive multimedia music technologies}.\hskip 1em plus 0.5em minus
  0.4em\relax IGI Global, 2008, pp. 50--79.

\bibitem{rebelo_optical_2012}
A.~Rebelo, I.~Fujinaga, F.~Paszkiewicz, A.~R. Marcal, C.~Guedes, and J.~d.~S.
  Cardoso, ``Optical music recognition: state-of-the-art and open issues,''
  \emph{International Journal of Multimedia Information Retrieval}, vol.~1,
  no.~3, pp. 173--190, 2012.

\bibitem{fujinaga2004staff}
I.~Fujinaga, ``Staff detection and removal,'' in \emph{Visual Perception of
  Music Notation: On-Line and Off Line Recognition}.\hskip 1em plus 0.5em minus
  0.4em\relax IGI Global, 2004, pp. 1--39.

\bibitem{fornes2008writer}
A.~Forn{\'{e}}s, J.~Llad{\'{o}}s, G.~S{\'{a}}nchez, and H.~Bunke, ``Writer
  identification in old handwritten music scores,'' in \emph{8th International
  Workshop on Document Analysis Systems}, Nara, Japan, 2008, pp. 347--353.

\bibitem{ridler1978picture}
T.~Ridler, S.~Calvard \emph{et~al.}, ``Picture thresholding using an iterative
  selection method,'' \emph{IEEE transactions on Systems, Man and Cybernetics},
  vol.~8, no.~8, pp. 630--632, 1978.

\bibitem{ballard1981generalizing}
D.~H. Ballard, ``Generalizing the hough transform to detect arbitrary shapes,''
  \emph{Pattern recognition}, vol.~13, no.~2, pp. 111--122, 1981.

\bibitem{Bainbridge2006}
D.~Bainbridge and T.~Bell, ``Identifying music documents in a collection of
  images,'' in \emph{7th International Conference on Music Information
  Retrieval}, Victoria, Canada, 2006, pp. 47--52.

\bibitem{dos_santos_cardoso_staff_2009}
J.~d.~S. Cardoso, A.~Capela, A.~Rebelo, C.~Guedes, and J.~Pinto~da Costa,
  ``Staff detection with stable paths,'' \emph{IEEE Transactions on Pattern
  Analysis and Machine Intelligence}, vol.~31, no.~6, pp. 1134--1139, 2009.

\bibitem{dalitz_comparative_2008}
C.~Dalitz, M.~Droettboom, B.~Pranzas, and I.~Fujinaga, ``A comparative study of
  staff removal algorithms,'' \emph{IEEE Transactions on Pattern Analysis and
  Machine Intelligence}, vol.~30, no.~5, pp. 753--766, 2008.

\bibitem{mahoney1982automatic}
J.~V. Mahoney, ``Automatic analysis of music score images,'' Ph.D.
  dissertation, Massachusetts Institute of Technology, Department of Electrical
  Engineering, 1982.

\bibitem{prerau-grammar}
D.~S. Prerau, ``Computer pattern recognition of standard engraved music
  notation,'' Ph.D. dissertation, Massachusetts Institute of Technology, 1970.

\bibitem{incremental-apacha}
A.~Pacha, ``Incremental supervised staff detection,'' in \emph{2nd
  International Workshop on Reading Music Systems}, J.~Calvo-Zaragoza and
  A.~Pacha, Eds., Delft, The Netherlands, 2019, pp. 16--20.

\bibitem{choudhury2000optical}
G.~S. Choudhury, M.~Droetboom, T.~DiLauro, I.~Fujinaga, and B.~Harrington,
  ``Optical music recognition system within a large-scale digitization
  project,'' in \emph{1st International Symposium on Music Information
  Retrieval}, 2000.

\bibitem{miyao1996note}
H.~Miyao and Y.~Nakano, ``Note symbol extraction for printed piano scores using
  neural networks,'' \emph{IEICE Transactions on Information and Systems}, vol.
  E79-D, no.~5, pp. 548--554, 1996.

\bibitem{pacha_learning_2019}
A.~Pacha, J.~Calvo-Zaragoza, and J.~Haji{\v{c}}~jr., ``Learning notation graph
  construction for full-pipeline optical music recognition,'' in \emph{20th
  International Society for Music Information Retrieval Conference}, 2019, pp.
  75--82.

\bibitem{mensural_pacha}
A.~Pacha and J.~Calvo-Zaragoza, ``Optical music recognition in mensural
  notation with region-based convolutional neural networks,'' in \emph{19th
  International Society for Music Information Retrieval Conference}, Paris,
  France, 2018, pp. 240--247.

\bibitem{baro2017optical}
A.~Bar{\'{o}}-Mas, ``Optical music recognition by long short-term memory
  recurrent neural networks,'' Master's thesis, Universitat Aut{\`{o}}noma de
  Barcelona, 2017.

\bibitem{droettboom2002optical}
M.~Droettboom, I.~Fujinaga, and K.~MacMillan, ``Optical music interpretation,''
  in \emph{Structural, Syntactic, and Statistical Pattern Recognition},
  T.~Caelli, A.~Amin, R.~P.~W. Duin, D.~de~Ridder, and M.~Kamel, Eds.\hskip 1em
  plus 0.5em minus 0.4em\relax Berlin, Heidelberg: Springer Berlin Heidelberg,
  2002, pp. 378--387.

\bibitem{calvo2019selectional}
J.~Calvo-Zaragoza and A.-J. Gallego, ``A selectional auto-encoder approach for
  document image binarization,'' \emph{Pattern Recognition}, vol.~86, pp.
  37--47, 2018.

\bibitem{gallego2017staff}
A.-J. Gallego and J.~Calvo-Zaragoza, ``Staff-line removal with selectional
  auto-encoders,'' \emph{Expert Systems with Applications}, vol.~89, pp.
  138--148, 2017.

\bibitem{masci2011stacked}
J.~Masci, U.~Meier, D.~Cire{\c{s}}an, and J.~Schmidhuber, ``Stacked
  convolutional auto-encoders for hierarchical feature extraction,'' in
  \emph{International conference on artificial neural networks}.\hskip 1em plus
  0.5em minus 0.4em\relax Springer, 2011, pp. 52--59.

\bibitem{blostein1992critical}
D.~Blostein and H.~S. Baird, ``A critical survey of music image analysis,'' in
  \emph{Structured Document Image Analysis}.\hskip 1em plus 0.5em minus
  0.4em\relax Springer Berlin Heidelberg, 1992, pp. 405--434.

\bibitem{dijkstra1959note}
E.~W. Dijkstra, ``A note on two problems in connexion with graphs,''
  \emph{Numerische mathematik}, vol.~1, no.~1, pp. 269--271, 1959.

\bibitem{iliescu1996proposed}
S.~Iliescu, R.~Shinghal, and R.~Y.-M. Teo, ``Proposed heuristic procedures to
  preprocess character patterns using line adjacency graphs,'' \emph{Pattern
  recognition}, vol.~29, no.~6, pp. 951--975, 1996.

\bibitem{tsukiyama1986method}
T.~Tsukiyama, Y.~Kondo, K.~Kakuse, S.~Saba, S.~Ozaki, and K.~Itoh, ``Method and
  system for data compression and restoration,'' Apr.~29 1986, uS Patent
  4,586,027.

\bibitem{carter1992automatic}
N.~P. Carter and R.~A. Bacon, ``Automatic recognition of printed music,'' in
  \emph{Structured Document Image Analysis}.\hskip 1em plus 0.5em minus
  0.4em\relax Berlin, Heidelberg: Springer Berlin Heidelberg, 1992, pp.
  456--465.

\bibitem{pugin2006optical}
L.~Pugin, ``Optical music recognitoin of early typographic prints using hidden
  {Markov} models,'' in \emph{7th International Conference on Music Information
  Retrieval}, Victoria, Canada, 2006, pp. 53--56.

\bibitem{calvo2016early}
J.~Calvo-Zaragoza, A.~H. Toselli, and E.~Vidal, ``Early handwritten music
  recognition with hidden markov models,'' in \emph{15th International
  Conference on Frontiers in Handwriting Recognition}.\hskip 1em plus 0.5em
  minus 0.4em\relax Institute of Electrical and Electronics Engineers Inc.,
  2017, pp. 319--324.

\bibitem{carter1989automatic}
N.~P. Carter, ``Automatic recognition of printed music in the context of
  electronic publishing.'' Ph.D. dissertation, University of Surrey (United
  Kingdom)., 1989.

\bibitem{rebelo2010optical}
A.~Rebelo, G.~Capela, and J.~d.~S. Cardoso, ``Optical recognition of music
  symbols,'' \emph{International Journal on Document Analysis and Recognition},
  vol.~13, no.~1, pp. 19--31, 2010.

\bibitem{ren2015faster}
S.~Ren, K.~He, R.~Girshick, and J.~Sun, ``Faster {R-CNN}: Towards real-time
  object detection with region proposal networks,'' in \emph{Advances in Neural
  Information Processing Systems 28}, C.~Cortes, N.~D. Lawrence, D.~D. Lee,
  M.~Sugiyama, and R.~Garnett, Eds.\hskip 1em plus 0.5em minus 0.4em\relax
  Curran Associates, Inc., 2015, pp. 91--99.

\bibitem{hajic2018towards}
J.~Haji{\v{c}}~jr., M.~Dorfer, G.~Widmer, and P.~Pecina, ``Towards
  full-pipeline handwritten {OMR} with musical symbol detection by u-nets,'' in
  \emph{19th International Society for Music Information Retrieval Conference},
  Paris, France, 2018, pp. 225--232.

\bibitem{dai2016r}
J.~Dai, Y.~Li, K.~He, and J.~Sun, ``R-fcn: Object detection via region-based
  fully convolutional networks,'' in \emph{Advances in neural information
  processing systems}, 2016, pp. 379--387.

\bibitem{liu2016ssd}
W.~Liu, D.~Anguelov, D.~Erhan, C.~Szegedy, S.~Reed, C.-Y. Fu, and A.~C. Berg,
  ``{SSD}: Single shot multibox detector,'' in \emph{Computer Vision -- ECCV
  2016}, B.~Leibe, J.~Matas, N.~Sebe, and M.~Welling, Eds.\hskip 1em plus 0.5em
  minus 0.4em\relax Cham: Springer International Publishing, 2016, pp. 21--37.

\bibitem{rothstein1995midi}
J.~Rothstein, \emph{MIDI: A comprehensive introduction}.\hskip 1em plus 0.5em
  minus 0.4em\relax AR Editions, Inc., 1995, vol.~7.

\bibitem{good2001musicxml}
M.~Good, ``Musicxml for notation and analysis,'' \emph{The virtual score:
  representation, retrieval, restoration}, vol.~12, pp. 113--124, 2001.

\bibitem{good2006lessons}
M.~Good and L.~Recordare, ``Lessons from the adoption of musicxml as an
  interchange standard,'' in \emph{Proceedings of XML}, 2006, pp. 5--7.

\bibitem{roland2002music}
P.~Roland, ``The music encoding initiative ({MEI}),'' in \emph{1st
  International Conference on Musical Applications Using XML}, 2002, pp.
  55--59.

\bibitem{hankinson2011music}
A.~Hankinson, P.~Roland, and I.~Fujinaga, ``The music encoding initiative as a
  document-encoding framework,'' in \emph{12th International Society for Music
  Information Retrieval Conference}, 2011, pp. 293--298.

\bibitem{jones2017musicowl}
J.~Jones, D.~de~Siqueira~Braga, K.~Tertuliano, and T.~Kauppinen, ``Musicowl:
  the music score ontology,'' in \emph{Proceedings of the International
  Conference on Web Intelligence}, 2017, pp. 1222--1229.

\bibitem{fazekas2010an}
G.~Fazekas, Y.~Raimond, K.~Jakobson, and M.~Sandler, ``An overview of
  {Semantic} {Web} activities in the {OMRAS}2 {Project},'' \emph{{Journal} of
  {New} {Music} {Research} special issue on {Music} {Informatics} and the
  {OMRAS}2 {Project}}, vol.~39, no.~4, pp. 295--311, 2011.

\bibitem{allik2016ontology}
A.~Allik, G.~Fazekas, and M.~Sandler, ``An ontology for audio features,'' in
  \emph{Proceedings of the International Society for Music Information
  Retrieval Conference}, 2016, pp. 73--79.

\bibitem{turchet2019cloud}
L.~Turchet, J.~Pauwels, C.~Fischione, and G.~Fazekas, ``Cloud-smart musical
  instrument interactions: Querying a large music collection with a smart
  guitar,'' \emph{ACM Transactions on the Internet of Things (In Press)}, 2020.

\bibitem{calvo2014recognition}
J.~Calvo-Zaragoza and J.~Oncina, ``Recognition of pen-based music notation: The
  {HOMUS} dataset,'' in \emph{22nd International Conference on Pattern
  Recognition}.\hskip 1em plus 0.5em minus 0.4em\relax Institute of Electrical
  \& Electronics Engineers (IEEE), 2014, pp. 3038--3043.

\bibitem{fornes2012cvc}
A.~Forn{\'{e}}s, A.~Dutta, A.~Gordo, and J.~Llad{\'{o}}s, ``{CVC-MUSCIMA}: A
  ground-truth of handwritten music score images for writer identification and
  staff removal,'' \emph{International Journal on Document Analysis and
  Recognition}, vol.~15, no.~3, pp. 243--251, 2012.

\bibitem{tuggener2018deepscores}
L.~Tuggener, I.~Elezi, J.~Schmidhuber, M.~Pelillo, and T.~Stadelmann,
  ``Deepscores - a dataset for segmentation, detection and classification of
  tiny objects,'' in \emph{24th International Conference on Pattern
  Recognition}, Beijing, China, 2018.

\bibitem{dorico}
Steinberg, ``{Dorico},'' \url{ https://new.steinberg.net/dorico/} (Jan. 2020).

\bibitem{rosegarden}
C.~Cannam, R.~Bown, and G.~Laurent, ``{Rosegarden},'' \url{
  https://www.rosegardenmusic.com/} (Jan. 2020).

\bibitem{novotny2015introduction}
J.~Novotn\`{y} and J.~Pokorn\`{y}, ``Introduction to optical music recognition:
  Overview and practical challenges,'' in \emph{Annual International Workshop
  on DAtabases, TExts, Specifications and Objects}, P.~J. Necasky~M.,
  Moravec~P., Ed.\hskip 1em plus 0.5em minus 0.4em\relax CEUR-WS, 2015, pp.
  65--76.

\end{thebibliography}

\end{document}